\documentclass[conference]{IEEEtran}
\IEEEoverridecommandlockouts
\usepackage{cite}
\usepackage{amsmath,amssymb,amsfonts}
\usepackage{algorithmic}
\usepackage{graphicx}
\usepackage{textcomp}
\usepackage{xcolor}
\def\BibTeX{{\rm B\kern-.05em{\sc i\kern-.025em b}\kern-.08em
    T\kern-.1667em\lower.7ex\hbox{E}\kern-.125emX}}
\begin{document}

\title{Hardware-Friendly Static Quantization Method for Video Diffusion Transformers}

\author{\IEEEauthorblockN{Sanghyun Yi}
\IEEEauthorblockA{\textit{Division of the Humanities and Social Sciences} \\
\textit{California Institute of Technology}\\
Pasadena, CA \\
syi@caltech.edu }
\and
\IEEEauthorblockN{Qingfeng Liu}
\IEEEauthorblockA{\textit{Device Solutions Research America} \\
\textit{Samsung Semiconductor Inc.}\\
San Diego, CA \\
qf.liu@samsung.com}
\and
\IEEEauthorblockN{Mostafa El-Khamy}
\IEEEauthorblockA{\textit{Device Solutions Research America} \\
\textit{Samsung Semiconductor Inc.}\\
San Diego, CA \\
m\_elkhamy@ieee.org}}
\maketitle

\begin{abstract}
Diffusion Transformers for video generation have gained significant
research interest since the impressive performance of SORA. Efficient deployment of such generative-AI models on GPUs has been demonstrated with dynamic quantization. 
However, resource-constrained devices cannot support dynamic quantization, and need static quantization of the models for their efficient deployment on AI processors.
In this paper, we propose a novel method for the post-training quantization of OpenSora\cite{opensora}, a Video Diffusion Transformer, without relying on dynamic quantization techniques. Our approach employs static quantization, achieving video quality comparable to FP16 and dynamically quantized ViDiT-Q methods, as measured by CLIP and VQA metrics. 
In particular, we utilize per-step calibration data to adequately provide a post-training statically quantized model for each time step, incorporating channel-wise quantization for weights and tensor-wise quantization for activations. 
By further applying the smooth-quantization technique, we can obtain high-quality video outputs with the statically quantized models. 
Extensive experimental results demonstrate that static quantization can be a viable alternative to dynamic quantization for video diffusion transformers, offering a more efficient approach without sacrificing performance.
\end{abstract}


\section{Introduction}
\label{sec:intro}
The use of generative AI for video creation, especially through diffusion techniques, has gained traction for various applications, ranging from content creation to complex simulation environments. As diffusion models become more common, enhancing their efficiency and reducing size for use in limited-resource settings becomes essential. One commonly employed approach for neural network compression is post-training quantization (PTQ). This method allows for reducing model size and improving inference speed without requiring model retraining, making it particularly attractive for large and computation-intensive models.
PTQ has been effective in optimizing large language models (LLMs) by reducing bit-width while keeping the performance of floating-point models \cite{gholami2021surveyquantizationmethodsefficient,zhu2024surveymodelcompressionlarge}. 

Despite such progress, quantization for video diffusion transformers (DiT) \cite{opensora} remains challenging due to their complexity. A recent work, Video Difussion Transformer Quantization (ViDiT-Q), suggests using dynamic quantization for these models \cite{zhao2024vidit}. However, this approach requires calculating the quantization parameters online during inference, which complicates hardware-level optimization, such as those for NPUs and Mobile System on Chip (SoC) \cite{park2022multi}.

To address this challenge, we propose static quantization for DiT models, calculating quantization parameters only during a calibration phase for use in inference. Our method includes aggregated static quantization that aggregates activation from all denoising steps to estimate a single set of quantization parameters that is used across denoising steps, and Time-Step-Wise (TSW) static quantization that estimates quantization parameters for each time step. Both approaches involve Channel-Wise (CW) quantization for weights, Tensor-Wise (TW) for activations, Aggregated or Time-step-wise Smooth Quantization (ASQ or TSQ).
Our static approaches match the performance of the original FP16 STDiT model (Spatial-Temporal DiT or OpenSora \cite{opensora}) and the dynamic quantization method from ViDiT-Q \cite{zhao2024vidit} in W8A8 quantization, with robust results at various precision levels.

\begin{figure}[t]
	\centering
	\includegraphics[width=1\linewidth]{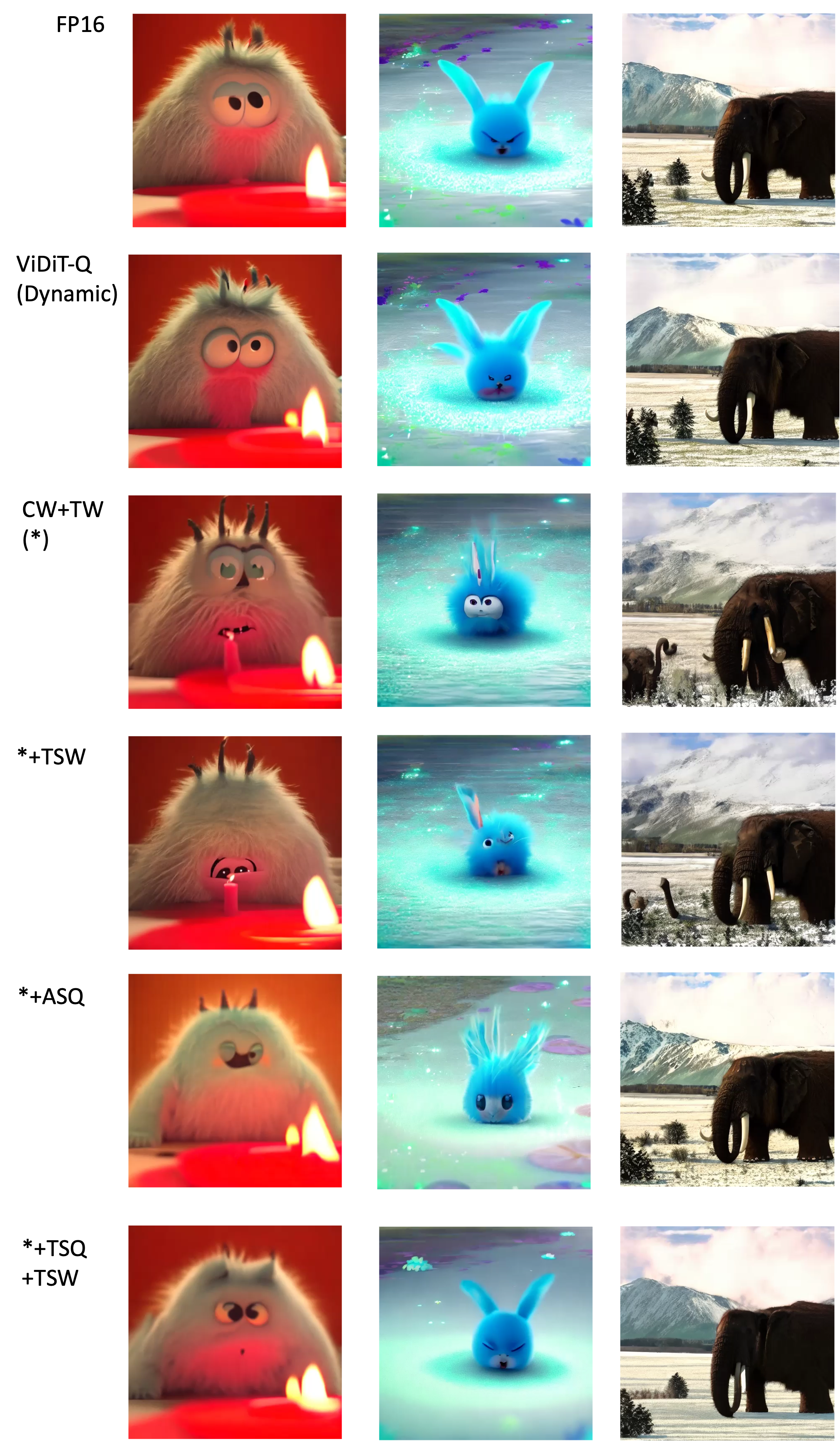}
	\caption{Example videos from various quantization methods
	}
	\label{fig1}
\end{figure}

For example, as shown in Figure~\ref{fig1}, our aggregated static quantization solution using CW, TW and ASQ (*+ASQ) showed comparable visual quality with dynamic quantization and the FP16 model. Our other proposal, TSW static quantization method using CW, TW and TSQ (*+TSQ+TSW) showed best alignment between prompt and generated videos.

\section{Related Works}
\label{sec:rw}
\textbf{PTQ  of Transformer Models:}
Numerous PTQ methods have been developed to meet the demands of transformer-based language models in industry \cite{zhao2023surveylargelanguagemodels}. For example, Activation-aware Weight Quantization (AWQ) quantizes weights based on the saliency of the corresponding activations and scales the weights per channel to minimize quantization errors \cite{lin2024awqactivationawareweightquantization}. Other methods also take into account the activation outliers in quantizing weights \cite{dettmers2023spqrsparsequantizedrepresentationnearlossless, lee2024owqoutlierawareweightquantization}.  Advancements in LLM weight quantization have even reached sub-2-bit quantization, utilizing ternary numbers for weights \cite{wang2023bitnetscaling1bittransformers, ma2024era1bitllmslarge}. Additionally, various methods have been proposed for PTQ of both weights and activations, with the primary challenge being the diverse distribution across activation channels. For instance, Smooth Quantization (SQ) smooths the weights and activation by scaling the channels, and Reorder-based Post-training Quantization (RPTQ) cluster the channels of weights and activations based on the similarity and quantizes per cluster \cite{xiao2024smoothquantaccurateefficientposttraining, yuan2023rptqreorderbasedposttrainingquantization}. 

A series of PTQ methods for Vision Transformer (ViT) was also suggested such as using mixed precision, attention weight order preserving quantization, and using multiple uniform quantization or log transformations to manage non-Gaussian activation distribution, \cite{liu2021posttraining, yuan2024ptq4vitposttrainingquantizationvision, lin2023fqvitposttrainingquantizationfully}.

\textbf{PTQ of Diffusion Models:}
However, the recurrent nature of diffusion models, which rely on the output of previous steps to sample the input for subsequent denoising steps, have limited the direct application of PTQ methods from other domains to diffusion model quantization. A key obstacle is the significant variation in activation distribution across time steps \cite{shang2023posttrainingquantizationdiffusionmodels}. Early work in PTQ for diffusion models mitigated this issue by generating calibration sets from the denoising process, with a focus on sampling across time steps to preserve output quality \cite{shang2023posttrainingquantizationdiffusionmodels, li2023qdiffusionquantizingdiffusionmodels,liu2024enhanceddistributionalignmentposttraining}. Other approaches involved using a neural network module to determine the time intervals for quantization and predict quantization parameters per interval or having different precision for different time steps and progressive calibration \cite{so2023temporaldynamicquantizationdiffusion, he2023ptqdaccurateposttrainingquantization,tang2024posttrainingquantizationtexttoimagediffusion}. Another study suggested a method using SQ, mixed precision across layers and focusing on the final time step in calibration for quantizing latent diffusion models \cite{yang2023efficientquantizationstrategieslatent}.

\textbf{PTQ of Diffusion Transformer Models:}
Recent advancements in video generation have increasingly adopted transformer models in denoising, such as Sora by OpenAI \cite{videoworldsimulators2024}. However, effective methods for quantizing Diffusion Transformer (DiT) models remain limited due to their unique characteristics compared to LLMs and CNN-based diffusion models. For example, unlike LLMs, DiT models experience significant variation across tokens, making the direct application of LLM quantization methods challenging \cite{zhu2024surveymodelcompressionlarge, zhao2024vidit}. Additionally, the activation distribution in DiT models varies significantly across time steps and between forward paths with or without conditional information from prompts \cite{zhao2024vidit}. The ViDiT-Q framework introduced techniques for quantizing DiT models, utilizing both path-wise dynamic quantization based on prompts and token-wise dynamic quantization for activations \cite{zhao2024vidit}. Despite its advantages, the framework requires real-time calculation of quantization parameters during inference, which complicates hardware optimizations, particularly for NPUs. Our methods avoid this problem by using static quantization while achieving the performance of the original model.

%To address this challenge, we propose a method to statically quantize DiT models by calculating quantization parameters during a calibration phase and using these fixed parameters during inference. Our method involves channel-wise quantization for weights, tensor-wise quantization for activations, smooth quantization, and estimating quantization parameters for each time step. Using this static quantization approach, we achieved performance comparable to the original FP16 model and the dynamic quantization method from ViDiT-Q \cite{zhao2024vidit} in W8A8 quantization for the STDiT (OpenSora \cite{opensora}) model.

\begin{figure*}[t]
	\centering
	\includegraphics[width=1\linewidth]{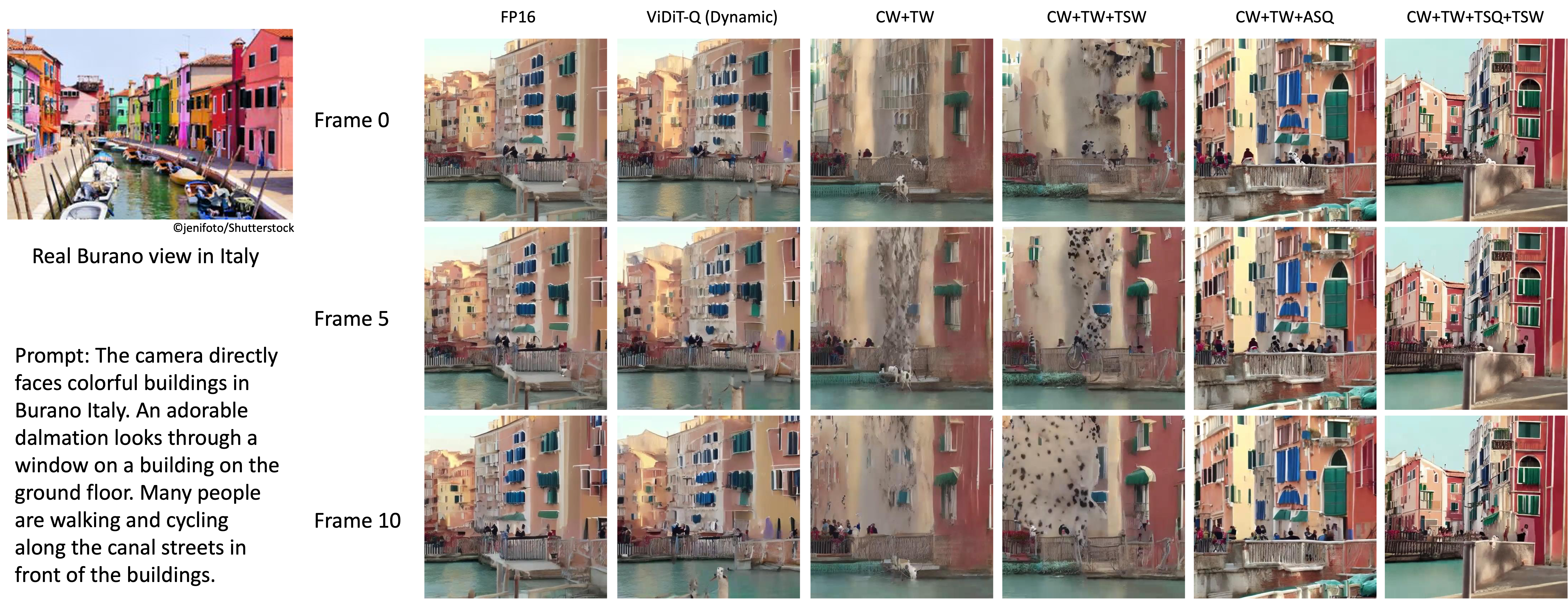}
	\caption{Example frames from generated videos for each quantization method and the corresponding prompt.
	}
	\label{fig_frames}
\end{figure*}

\section{Our Method}

\begin{figure}[!htb]
	%\centering
	\includegraphics[width=1\linewidth]{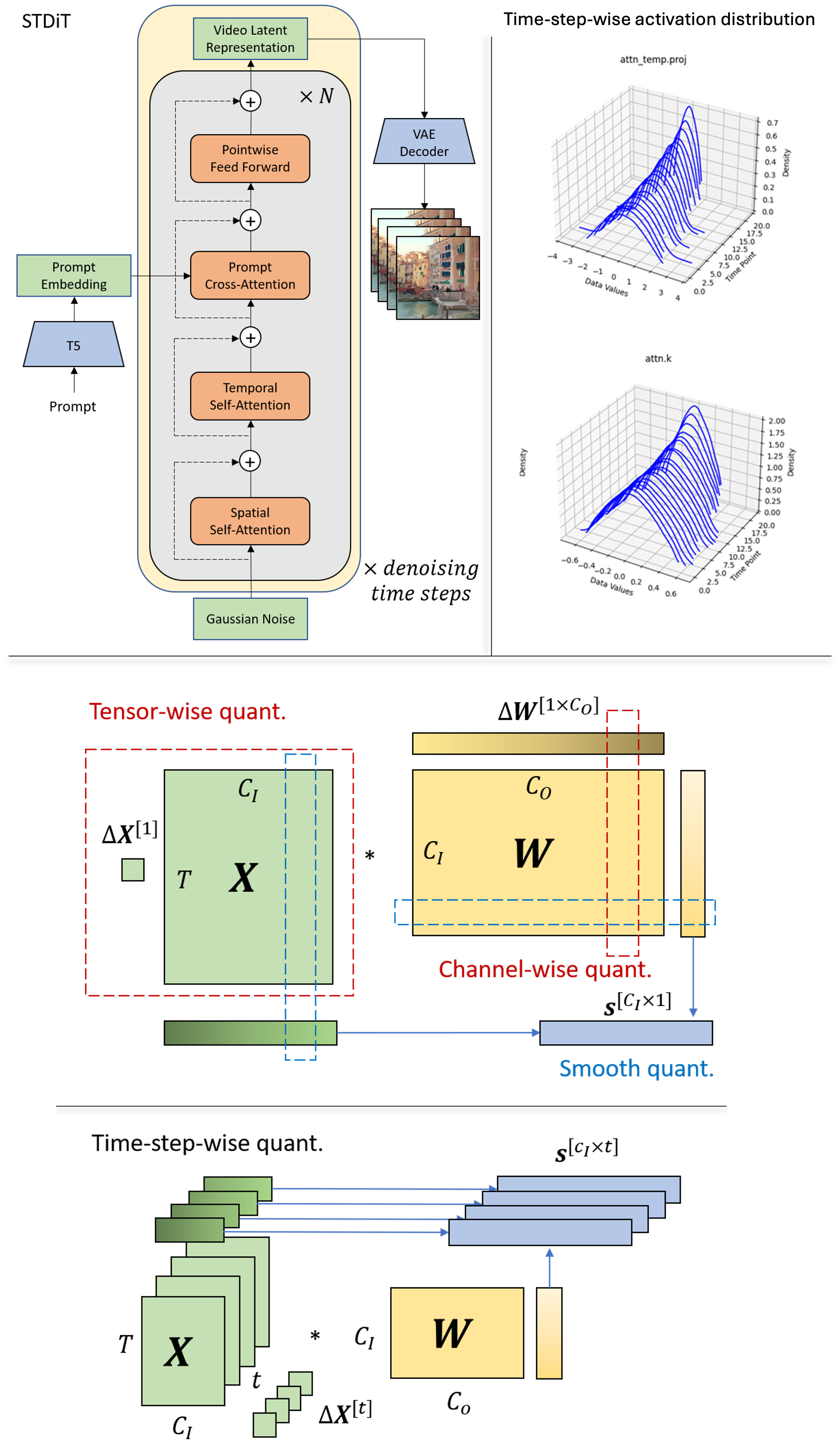}
	\caption{\textbf{Overview of the proposed method} Upper left figure shows the architecture of STDiT. The upper right figure shows the change of activation distribution  across time steps from STDiT. Lower figures summarized the components of our method: Channel-Wise (CW) quantization for weights, Tensor-Wise (TW) quantization for activations, Aggregated or Time-step-wise Smooth Quantization (ASQ, TSQ), and Time-Step-Wise (TSW) static quantization for each diffusion step. The linear layers in the attention and feed forward layers in the transformer blocks (orange-colored) of STDiT were quantized. $N$ is the number of transformer blocks (28 in STDiT v1.0). $\it{\bf{X}}$ is the activation and $\it{\bf{W}}$ is the weight matrix where $T$ is the number of tokens, $C_I$ and $C_O$ is the input and output channel sizes and $t$ is the number of denoising time steps.
	}
	\label{fig0}
\end{figure}

The proposed methods, aggregated static quantization and time-step-wise static quantization consist of three different components to quantize the transformer denoiser of DiT models: Channel-Wise (CW) weight quantization, Tensor-Wise (TW) activation quantization and Aggregated of Time-step-wise Smooth Quantization (ASQ or TSQ). Each component's will be explained as Aggregated Static Quantization version first and the Time-step-wise Static Quantization version will be discussed in the next subsection. Using these methods, linear layers of the spatial self-attention, temporal self-attention, prompt cross-attention, and pointwise feed forward layers in the transformer blocks were quantized (See the upper left plot of Figure \ref{fig0}).

\subsection{Aggregated Static Quantization}
The first component of our method involves Channel-Wise (CW) quantization of the weight matrix to mitigate quantization errors arising from channel-wise variance \cite{zhao2024vidit}. The minimum and maximum values for each channel in weights were extracted and stored. Then the minimum and maximum were used to calculate the bin sizes ($\Delta W_i$) and zero points ($z_{W_i}$) channel-wise using the following equation \ref{eq1}.
\begin{equation}\label{eq1}
\begin{split}
\Delta W_i = \frac{\max(W_i)-\min(W_i)}{2^b}, \quad
z_{W_i} = \frac{\min(W_i)}{\Delta W_i},\\ 
b\text{ is the bit width and } i\text{ is the channel index.}
\end{split}
\end{equation}
Thus the bin sizes and zero points for CW quantization of a weight matrix have $[1\times C_O]$ dimensions where $C_O$ is the output channel size (see the middle row of Figure \ref{fig0}). 

The second component is Tensor-Wise (TW) quantization of the activation matrix. While dynamic token-wise quantization is widely used for transformer models, it is not feasible to estimate statistics to cover the variance of each token activation during inference in a static manner due to the heterogeneity across inference samples. Instead, the simplest method, which involves estimating the minimum and maximum values of activations tensor-wise, was utilized. Therefore, the bin sizes ($\Delta X$) and zero points ($z_X$) for TW quantization of a activation matrix are scalar values (see the middle row of Figure \ref{fig0}).

Next, as the third component, we explored the variants of smooth quantization method to migrate the quantization difficulty from activations to weights \cite{xiao2024smoothquantaccurateefficientposttraining}. 
In particular, we proposed the Aggregated Smooth Quantization (ASQ) to be applied in the aggregated static quantization method. 
In ASQ, the maximum absolute values of weights for each channel were extracted ($\max(|W_{i})|$) to calculate the smooth quantization scaling term ($s_{i}$) for each channel $i$. Then, the maximum absolute values of activations for each channel were averaged across the batch ($\max(|X_{i}|)$). The maximum absolute values of activations for the entire calibration set were then calculated using a running average with a momentum of 0.95, aggregating information across time steps. The final scaling term for smooth quantization was determined using equation \ref{eq2}, where $\alpha\in[0,1]$ is a hyperparameter that needs to be tuned. Therefore, in ASQ, a single scaling term was obtained for each channel, which was used for the entire denoising steps.

\begin{equation}\label{eq2}
\begin{split}
Y = \left(X \cdot \text{diag}(s)^{-1}\right) \cdot \left(\text{diag}(s) \cdot W\right) = \hat{X} \cdot \hat{W} \\
s_{i} = \frac{\max(|X_{i}|)^\alpha}{\max(|W_{i}|)^{1-\alpha}}, \quad i\text{ is the channel index}
\end{split}
\end{equation}

As a result, the scaling term ($s_{i}$) has the size of $[C_I\times 1]$ where the $C_I$ is the input channel size (see the middle row of Figure \ref{fig0}). When the ASQ was applied, CW weight quantization and TW activation quantization was applied to the smoothed $\hat{X}$ and $\hat{W}$, instead of the raw $X$ and $W$. Therefore, applying ASQ and doing TW activation quantization can be viewed as a method that handles channel-wise variance of the activation without doing CW quantization on the activation matrix.

\subsection{Time-step-wise Static Quantization}

One of our main contributions is a method that estimates quantization parameters for individual denoising time steps during the calibration stage only, enabling Time-Step-Wise (TSW) static quantization. Unlike previously suggested dynamic quantization method that estimates quantization parameters for each denoising time step during the inference \cite{zhao2024vidit}, our TSW static quantization estimates the parameters during the calibration stage only, and use the estimated parameters without any updates for the inference, facilitating efficient hardware-level optimization.

Thus TSW static quantization can handle the time-step-wise variance in activation distributions (See the upper right plot in Figure \ref{fig0}) without additional calculation during the inference. The essence of implementing the static TSW quantization is to use the calibration set to estimate the parameter statistics for each denoising step. 

When the TSW static quantization is applied, the bin sizes ($\Delta X$) and zero points ($z_X$) for the TW quantization have $[1\times t]$ dimensions. The smooth quantization scaling term ($s_i$) has the $[C_I \times t]$ dimension where $t$ is the number of denoising time steps, which we called Time-step-wise Smooth Quantization, or TSQ. The CW quantization stays same as it quantizes weight not activations. See the last row of Figure \ref{fig0}). During inference, these fixed, pre-estimated statistics are used without further calculations or updates.

TSW static quantization can also be applied in a coarser manner, where quantization parameters are estimated not for each time step but across time ranges that group multiple denoising steps. For instance, estimating parameters over the entire denoising process corresponds to a single time-range static quantization. Dividing the process into two time ranges (e.g., early and late stages) and estimating parameters for each range leads to a two-time-range static quantization. The effects of different levels of granularity in the estimation of static quantization parameters are discussed further in Section \ref{tsw}.

\begin{table}[!htb]
	%\centering
	\caption{Quantization methods on the open-sora prompt set}
	\label{tab1}
    \small % Makes the font size smaller
    \setlength{\tabcolsep}{2pt} % Reduce padding between columns
	\begin{tabular}{l | l | c | c | c | c %| c | c
	}
		Method & Bit (W/A) & CLIPSIM & CLIP-temp & VQA-a & VQA-t %& IQA-a & IQA-t
		\\
		\hline
		FP16 \cite{opensora}& 16/16 & 0.1950 & 0.9982 & 54.2730 & 49.9179 %& 4.8132 & 4.6001
		\\
        Dynamic \cite{zhao2024vidit} & 8/8 & 0.1960 & 0.9982 & 53.3998 & 49.4671 %& 4.8219 & 4.5962
        \\
        \hline
        CW+TW (*) & 8/8 & 0.1931 & 0.9966 & 44.2729 & 38.6906 %& 4.5347 & 4.3631
        \\
        *+TSW & 8/8 & 0.1931 & 0.9972 & 42.4221 & 40.8720 %& 4.6094 & 4.4304
        \\
        %*+LC+TSW & 8/8 & 0.1910 & 0.9986 & 14.2270 & 16.6479 %& 4.2340 & 4.6134
        %\\
        *+ASQ & 8/8 & 0.1926 & \bf0.9988 & \bf52.9969 & \bf49.9545 %& 4.8030 & 4.6242
        \\
        *+TSQ+TSW & 8/8 & \bf0.1967 & 0.9987 & 47.5200 & 39.0468 %& \bf4.8104 & \bf4.6884
        \\
		\hline
	\end{tabular}
\end{table}
\begin{table}[!htb]
\centering
	\caption{Quantization methods on UCF-101} 
    \small % Makes the font size smaller
    \setlength{\tabcolsep}{2pt} % Reduce padding between columns
	\label{tab2}
	\begin{tabular}{l | l | c | c | c | c %| c | c
	}
		Method & Bit (W/A) & CLIPSIM & CLIP-temp & VQA-a & VQA-t %& IQA-a & IQA-t
		\\
		\hline
		FP16 \cite{opensora}& 16/16 & 0.2058 & 0.9970 & 26.3483 & 34.5356 %& 4.5385 & 5.0782
		\\
        Dynamic\cite{zhao2024vidit} & 8/8 & 0.2049 & 0.9968 & 25.6435 & 34.6162 %& 4.5249 & 5.1175
        \\
        \hline
        CW+TW (*) & 8/8 & 0.2033 & 0.9955 & 18.1882 & 22.8154 %& 4.3827 & 4.8845
        \\
        *+TSW & 8/8 & 0.2025 & 0.9958 & 17.9768 & 24.7684 %& 4.4000 & 4.9351
        \\
        *+ASQ & 8/8 & 0.2048 & 0.9975 & \bf24.6499 & \bf31.9683 %& \bf4.5345 & \bf5.1199
        \\
        *+TSQ+TSW & 8/8 & \bf0.2084 & \bf0.9986 & 10.2957 & 17.4171 %& 4.3674 & 5.1182
        \\
		\hline
	\end{tabular}
\end{table}

\begin{table}[!htb]
	\caption{Robustness test of the proposed methods}
    \centering
    \small % Makes the font size smaller
    \setlength{\tabcolsep}{2pt} % Reduce padding between columns
	\label{tab3}
	\begin{tabular}{l | l | c | c | c | c %| c | c
	}
		Bit (W/A) & Method & CLIPSIM & CLIP-temp & VQA-a & VQA-t %& IQA-a & IQA-t
		\\
		\hline\hline
  FP16 \cite{opensora}& - & 0.1950 & 0.9982 & 54.2730 & 49.9179 %& 4.8132 & 4.6001
  \\
  \hline
    4/4  & *+TSQ+TSW & 0.1569 & 0.9967 & 0.0257 & 0.0008 %& 3.9653 & 4.6572
    \\
      & *+ASQ & 0.1538 & \bf0.9974 & 0.0171 & 0.0009 %& 4.0158 & 4.5555
      \\
      & Dynamic \cite{zhao2024vidit} & \bf0.1786 & 0.9729 & \bf2.7089 & \bf0.4849 \\
      \hline
    4/6  & *+TSQ+TSW & \bf0.1909 & 0.9981 & \bf4.8082 & \bf5.4387 %& 4.0776 & 4.1705
    \\
    & *+ASQ & 0.1838 & \bf0.9983 & 0.0557 & 0.0686 %& 3.8255 & 4.5854
    \\
    & Dynamic \cite{zhao2024vidit} & 0.1817 & 0.9976 & 0.1681 & 0.2958 \\
    \hline
    6/6  & *+TSQ+TSW & 0.1864 & 0.9981 & 9.2497 & 9.2459 %& 4.2499 & 4.1549
    \\
    & *+ASQ & 0.1829 & \bf0.9986 & 0.0961 & 0.0781 %& 3.8122 & 4.5334
    \\
    & Dynamic \cite{zhao2024vidit} & \bf0.1881 & 0.9981 & \bf18.9117 & \bf15.1275 \\
    \hline
    4/8  & *+TSQ+TSW & \bf0.1922 & \bf0.9994 & 24.5604 & \bf24.0194 %& 4.6170 & 4.5706
    \\
    & *+ASQ & 0.1849 & 0.9984 & \bf26.9354 & 15.8466 %& 4.5022 & 4.2895
    \\
    & Dynamic \cite{zhao2024vidit} & 0.1849 & 0.9986 & 0.0404 & 0.3072 \\
    \hline
    6/8  & *+TSQ+TSW & \bf0.1962 & 0.9987 & \bf36.0467 & \bf40.6366 %& 4.5335 & 4.6191
    \\
    & *+ASQ & 0.1910 & \bf0.9991 & 25.6601 & 17.4476 %& 4.4841 & 4.4779
    \\
    & Dynamic \cite{zhao2024vidit} & 0.1903 & 0.9990 & 28.2134 & 21.8238 \\
    \hline
    8/8  & *+TSQ+TSW & \bf0.1967 & 0.9987 & 47.5200 & 39.0468 %& 4.8104 & 4.6884
    \\
    & *+ASQ & 0.1926 & \bf0.9988 & \bf53.9969 & \bf49.9545 %& 4.8030 & 4.6242
    \\
    & Dynamic \cite{zhao2024vidit} & 0.1960 & 0.9982 & 53.3998 & 49.4671 \\
    \hline
    4/16 & *+TSQ+TSW & \bf0.1911 & \bf0.9996 & 28.3527 & 12.5568 %& 4.6650 & 4.4838
    \\
    & *+ASQ & 0.1895 & 0.9988 & \bf31.6746 & \bf35.4823 %& 4.4511 & 4.5720
    \\
    & Dynamic \cite{zhao2024vidit} & 0.1871 & 0.9972 & 0.0237 & 0.0240 \\
    \hline
    6/16 & *+TSQ+TSW & \bf0.1960 & \bf0.9996 & 26.2750 & 17.4801 %& 4.6757 & 4.5786
    \\
    & *+ASQ & 0.1959 & 0.9989 & \bf44.6240 & \bf38.0148 %& 4.6417 & 4.6155
    \\
    & Dynamic \cite{zhao2024vidit} & 0.1854 & 0.9983 & 8.3704 & 10.7042 \\
    \hline
    8/16 & *+TSQ+TSW & 0.1953 & \bf0.9988 & 46.8446 & 38.9204 %& 4.7489 & 4.6439
    \\
    & *+ASQ & \bf0.1963 & 0.9982 & \bf53.2460 & \bf51.9194 %& 4.8196 & 4.6157
    \\
    & Dynamic \cite{zhao2024vidit} & 0.1871 & 0.9981 & 24.0312 & 26.8858 \\
    \hline\hline
    Average & *+TSQ+TSW & \bf0.1891 & \bf0.9986 & 24.8537 & 20.8162 %& \bf4.4826 & 4.5074
    \\
    & *+ASQ & 0.1856 & 0.9985 & \bf26.1451 & \bf23.2014 %& 4.3728 & \bf4.5410
    \\
    & Dynamic \cite{zhao2024vidit} & 0.1866 & 0.9953 & 15.0964 & 13.9022 \\
		\hline
	\end{tabular}
\end{table}

%-------------------------------------------------------------------------
\begin{figure*}[t]
	\centering
	\includegraphics[width=1.0\linewidth]{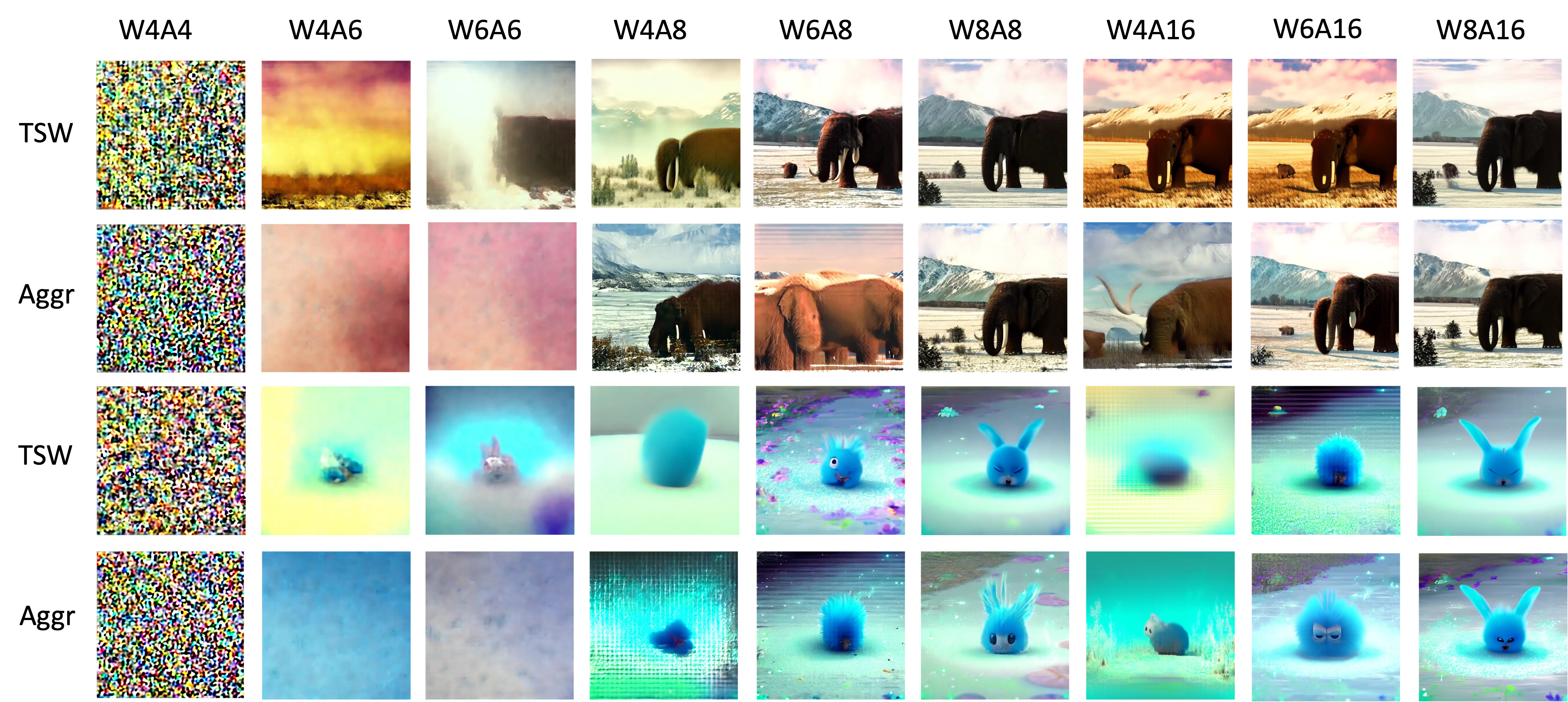}
	\caption{Robustness comparison between the CW+TW+ASQ (Aggr) and the CW+TW+TSQ+TSW (TSW). Additional examples in the supplemental material.}
	\label{fig2}
\end{figure*}

%\begin{figure*}[t]
%	\centering
%	\includegraphics[width=1\linewidth]{figures/3.png}
%	\caption{Robustness test examples from the CW+TW+ASQ model
%	}
%	\label{fig3}
%\end{figure*}

%------------------------------------------------------------------------
\begin{table}[!htb]
	\caption{Granular TSW quantization on the open-sora prompt set}
	\label{tab5}
    \centering
	\begin{tabular}{l | c | c | c | c %| c | c
	}
		Method & CLIPSIM & CLIP-temp & VQA-a & VQA-t %& IQA-a & IQA-t
		\\
		\hline
 *+ASQ (=1TR)  & 0.1926 & 0.9988 & 52.9969 & 49.9545 %& 4.8030 & 4.6242
 \\
    *+TSQ+2TR  & 0.1960 & 0.9981 & 42.2731 & 43.7837 %& 4.7081 & 4.4828
    \\
    *+TSQ+4TR  & 0.1957 & 0.9978 & 43.3432 & 46.1740 %& 4.6421 & 4.6014
    \\
    *+TSQ+10TR  & 0.1946 & 0.9978 & 40.5217 & 42.3603 %& 4.4233 & 5.0389
    \\
    *+TSQ+TSW (=20TR)  & 0.1967 & 0.9987 & 47.5200 & 39.0468 %& 4.8104 & 4.6884
    \\
		\hline
	\end{tabular}
\end{table}
\begin{table}[!htb]
	\caption{Granular TSW quantization on UCF-101}
	\label{tab6}
        \centering
	\begin{tabular}{l | c | c | c | c %| c | c
	}
		Method & CLIPSIM & CLIP-temp & VQA-a & VQA-t %& IQA-a & IQA-t
		\\
		\hline
 *+ASQ (=1TR)  & 0.2048 & 0.9975 & 24.6499 & 31.9683 %& 4.5345 & 5.1199
 \\
    *+TSQ+2TR  & 0.2039 & 0.9966 & 18.0531 & 23.2528 %& 4.5379 & 4.9975
    \\
    *+TSQ+4TR  & 0.2054 & 0.9967 & 18.3887 & 26.8416 %& 4.4233 & 5.0389
    \\
    *+TSQ+10TR  & 0.2063 & 0.9967 & 14.2008 & 22.1882 %& 4.4233 & 5.0389
    \\
    *+TSQ+TSW (=20TR)  & 0.2084 & 0.9986 & 10.2957 & 17.4171 %& 4.3674 & 5.1182
    \\
		\hline
	\end{tabular}
\end{table}

\section{Experiment}
\subsection{Implementation Details and Experimental Settings}
We evaluated the performance of our methods on the STDiT v1.0 using various bit-widths and evaluation settings \cite{opensora}. For quantization, we adopted the min-max quantization scheme, where the quantization parameters for actions were estimated using a calibration set. This calibration set was generated using example prompts to create a calibration dataset for ViDiT-Q, consisting of 10 prompts \cite{zhao2024vidit}. 
%We also tested larger calibration (LC) set selected from WebVid-10M which was used as a training set for STDiT \cite{Bain21}. About 600 prompts from WebVid-10M were embedded using CLIP-ViT-L14 and clustered into 600 groups using K-Means clustering \cite{radford2021learningtransferablevisualmodels}. Then the prompts closest to each centroid were selected as the calibration set. 
For evaluation, we utilized CLIPSIM and CLIP-temp, both scaled from 0 to 1, to measure text-video alignment and temporal semantic consistency of the generated videos \cite{wu2021godivageneratingopendomainvideos, esser2023structurecontentguidedvideosynthesis}. The CLIPSIM is the average CLIP cosine similarity between the prompt CLIP embedding and each frame's CLIP embedding. The CLIP-temp is calculated by averaging across CLIP cosine similarities between CLIP embeddings of consecutive two frames. To assess video quality, we used VQA-aesthetic and VQA-technical scores, scaled from 0 to 100, to evaluate high-level aesthetics and low-level technical soundness \cite{wu2023exploringvideoqualityassessment}. The VQA scores are automatically evaluated by neural networks trained on large video datasets, which were annotated by humans for aesthetic appeal and technical quality. For the evaluations, implementations in the EvalCrafter were used \cite{liu2024evalcrafterbenchmarkingevaluatinglarge}. %Additionally, we employed IQA-aesthetic and IQA-technical, scaled from 1 to 10, for image evaluation, supplementing the VQA metrics by calculating scores for each frame and averaging them across frames \cite{Talebi_2018}. 
The experiments were conducted on UCF-101 \cite{soomro2012ucf101dataset101human} and OpenSora\cite{opensora} datasets, on a single A100 GPU using 20 denoising steps and an IDDPM scheduler with a CFG scale of 7.0. The $\alpha$ parameter for SmoothQuant was grid-searched from 0.1 to 1.0 in 0.1 increments, chosen based on the CLIPSIM score. For dynamic quantization, $\alpha = 0.625$ (from ViDiT-Q) was used; for CW+TW+ASQ, $\alpha = 0.4$; and for CW+TW+TSQ+TSW, $\alpha = 0.2$. 

\subsection{Main Results and Ablation Studies}

First, we compared our STDiT quantization approaches (*+ASQ and *+TSQ+TSW) with previous methods (Tables \ref{tab1} and \ref{tab2}). Our methods showed superior performance across various metrics. Specifically, *+TSQ+TSW excelled in producing semantically aligned videos. For example, frames from the generated videos in Figure \ref{fig_frames} show that our *+TSQ+TSW method most accurately captured the vibrant characteristics of the city of Burano (see also Figure \ref{fig1}). The aggregated method, *+ASQ achieved video quality comparable to ViDiT-Q's dynamic quantization according to VQA metrics. 

Intuitively, it might be thought that aggregated static quantization would perform worse than TSW static quantization. However, we observed on-par or even superior performance from the *+ASQ models for a subset of metrics. This is attributed to the fact that when ASQ (also TSQ) is applied, within-a-channel activity variations are smoothed out, allowing the maximum absolute activation to be normalized by $max(|X_i|)^\alpha$, which ensures that the maximum absolute value of the smoothed activation is less than 1, bringing the range of activations within $[-1, 1]$ regardless of the denoising steps, and making TSW less necessary. 

In relation to this analysis, the ablation study further confirmed the importance of applying ASQ and TSQ for the best static quantization performance (Tables \ref{tab1} and \ref{tab2}).

However, while not using TSQ but using ASQ does not lead to significant failures—due to the activation distribution remaining within the range of the calibrated parameters—it can result in less precise quantization. This imprecision arises because the activation range for some time steps may only cover subset of $[-1, 1]$. Consequently, we can imagine *+ASQ is more sensitive to changes in bit-widths than *+TSQ+TSW, which will be discussed in the next section \ref{tsw}.

%However, using the larger calibration (LC) set didn't improve the performance much so we didn't explore this direction further. 

\subsection{Method Robustness Across Different Bit-Widths}\label{tsw}
We assessed the robustness of *+TSQ+TSW and *+ASQ across various bit-widths using the open-sora prompt set and compared them to the dynamic quantization used in ViDiT-Q for W8A8 quantization \cite{zhao2024vidit}. Table \ref{tab3} shows that both static quantization methods outperformed dynamic quantization across a broader bit-width range. The averaged score across bit-width in Table \ref{tab3} revealed that the *+TSQ+TSW model generated videos with better semantic alignment with the prompt and temporal consistency, while the *+ASQ model produced videos with more aesthetic appeal and less noise and distortion. 

Furthermore, qualitative analysis showed that *+TSQ+TSW generated recognizable videos in wider ranges of bit-width than the *+ASQ which confirming our analysis in the previous section (see Figure \ref{fig2}).

\subsection{Effect of Time-Step-Wise Quantization}
We further explored the impact of time-step-wise quantization by grouping time steps into several time ranges (TR), estimating quantization parameters separately for each group. For instance, the 2TR method grouped the time steps into two ranges and estimated quantization parameters for each range, while the 4TR method did so for four separate time ranges. Therefore, for example, the SQ scaling term will have $[C_i\times 2]$ or $[C_i\times 4]$ dimensions for 2TR or 4TR methods. As shown in Tables \ref{tab5} and \ref{tab6}, implementing time-step-wise quantization improved semantic adherence to the prompt, as measured by CLIP, but compromised video quality metrics (VQA-a and VQA-t) and the model size. This trade-off is evident in the qualitative analysis of the generated videos (see the last two columns of Figure \ref{fig1} and Figure \ref{fig2}).

\subsection{Cross-Dataset Evaluation}
Although calibration was performed using the OpenSora prompt set from the ViDiT-Q paper, we validated our model’s generalization ability by evaluating it on UCF-101. The results in Tables \ref{tab2} and \ref{tab6} demonstrate that our static quantization approach maintains competitive performance, validating its cross-dataset generalization capability.

\subsection{Temporal Efficiency}
We evaluated the execution time of different quantization methods by measuring the time taken to generate 48 videos using the OpenSora prompt set on an NVIDIA A100 GPU. As shown in Table \ref{tab7}, both CW+TW+ASQ and CW+TW+TSQ+TSW showed significantly reduced inference time compared to the dynamic quantization approach.

\begin{table}[t]
\centering
\caption{Execution time for generating 48 videos using OpenSora prompts on an A100 GPU}
\label{tab7}
    \begin{tabular}{c|c}
        \hline
        Method & Execution time (s)\\
        \hline
        Dynamic & 2259 \\
        CW+TW+ASQ & 1942 \\
        CW+TW+TSQ+TSW & 1919 \\
        \hline
    \end{tabular}
\end{table}

\section{Conclusion}

Modern Mobile SoC hardware lacks support for any form of dynamic quantization due to its inability to calculate activation or weight statistics at runtime \cite{park2022multi}. Our proposed static quantization schemes provide a viable solution for deployment on these devices, as they eliminate the need for online statistic computation, enabling simpler hardware design with lower power and area requirements.

This paper introduces two static quantization methods—aggregated and time-step-wise—that match the performance of previously proposed dynamic quantization for DiT models. By incorporating channel-wise weight quantization, tensor-wise activation quantization and smooth quantization, either aggregated or time-step-wise, our approach successfully maintains the performance of the original STDiT (Open-Sora) model.

The proposed methods could broaden the use of diffusion-based generative models in mobile AI devices, though limitations at narrow bit-widths suggest room for future enhancement. Additionally, assigning different quantized models for each time step means intelligent processing to swap models across time steps is needed to hide the latency in loading new model parameters. Balancing between fine-grained time step quantization and grouping time steps into ranges, while considering hardware constraints, will be crucial for optimizing the method's practical application.

%\subsection{Supplemental material}
%Please refer to the provided slides for videos of the figures.

\bibliographystyle{IEEEbib}
\bibliography{icme2025references}

\end{document}